Nick Williams 2014

Title: Conclusions from a NAÏVE Bayes Operator Predicting the Medicare 2011 Transaction Data Set

Abstract:

Introduction: The United States Federal Government operates one of the world's largest medical insurance programs, Medicare, to ensure payment for clinical services for the elderly, illegal aliens and those without the ability to pay for their care directly. The program is mired in controversy over its costs and consequences. This paper evaluates the Medicare 2011 Transaction Data Set which details the transfer of funds from Medicare to private and public clinical care facilities for specific clinical services (DRG) for the operational year 2011.

Methods: Data mining was conducted to establish the relationships between reported and computed transaction values in the data set to better understand the drivers of Medicare transactions at a programmatic level. Naïve Bayes is a classification algorithm for establishing the predictability of a labeled class (variable) of data given the presence of other data elements. This paper adapts the traditional Naïve effort to characterize the added model value that class attributes have on a data set permuted by redaction; one class redaction set per available class. Second the cross validation algorithm was trained to the redaction sets, computing model values for the highest accuracy and Kappa, lowest accuracy and Kappa as well as MIKRO or average accuracy and kappa of the cross trained models per redacted set.

Results: The models averaged 88% for average model accuracy and 38% for average Kappa during training. Some reported classes are highly independent from the available data as their predictability remains stable regardless of redaction of supporting and contradictory evidence. DRG or procedure type appears to be unpredictable from the available financial transaction values. This independence from the financial and geographic values in the data set across the redaction matrix suggests that charges, payments and financial losses are not driven by procedure, even when adjusted for discharge.

Conclusions: Overlay hypotheses such as charges being driven by the volume served or DRG being related to charges or payments is readily false in this analysis despite 28 million Americans being billed through Medicare in 2011 and the program distributing over 70 billion in this transaction set alone. It may be impossible to predict the dependencies and data structures the payer of last resort without data from payers of first and second resort. Political concerns about Medicare would be better served focusing on these first and second order payer systems as what Medicare costs is not dependent on Medicare itself.

Overview

The Medicare 2011 Transaction Data Set (MTDS) is a joined transformation of the Medicare Provider Charge Data Sets which aggregates and compiles all values from the Inpatient and Outpatient sets for the year 2011[i]. This MTDS details the transfer of money from the United States Federal Government's Medicare Insurance Program to provider facilities for specific clinical services. While the complexity of the multi-billion dollar program is as vast and complicated as the clinical experiences of the tens of millions it serves and the providers it enriches[ii] (or impoverishes[iii]), there has been little work to understand the structure and independence of financial transactions. Rather glib understandings and shallow evaluations have dogged the program from its inception, fueling various and sometimes contradictory political debates from the proper cost of medical care[iv], to the explosion of physician salaries relative to the incomes of the general public whom they claim to serve. Lately the 'immigration debate"[v] (Medicare will pay for clinical procedures for illegal aliens) has refueled the politicization of a largely unevaluated program, mirroring yesteryear's Medicare debate about the racial integration of public and private hospitals using federal funds following the civil rights act in 1964[vi]. What we want Medicare to prove and what Medicare transactions actually detail will remain a mystery as long as Medicare remains under described in the terms which it actually operates.

As the payer of last resort Medicare is a stop gap measure that insures facilities will be paid regardless of the financial constraints of whom they practice on. This utility, while essential for any market system that offers essential clinical care at cost and often for profit, has some glaring undersides including the inflation of the need for care[vii], the conflation of care and cost independent of price, as well as wild departures between the charge for identical care facility to facility within the same country and reimbursement system. As specialty care and unique clinical experiences of patients become increasingly legible to reimbursement schemes that aim to leave no one behind they fall under increasing weight to become economically responsible and confrontational to 'cost' without accounting for or documenting profits.

Graph One: MTDS Total Charges, Costs, Payment and Discharges by Facility DRG Transaction

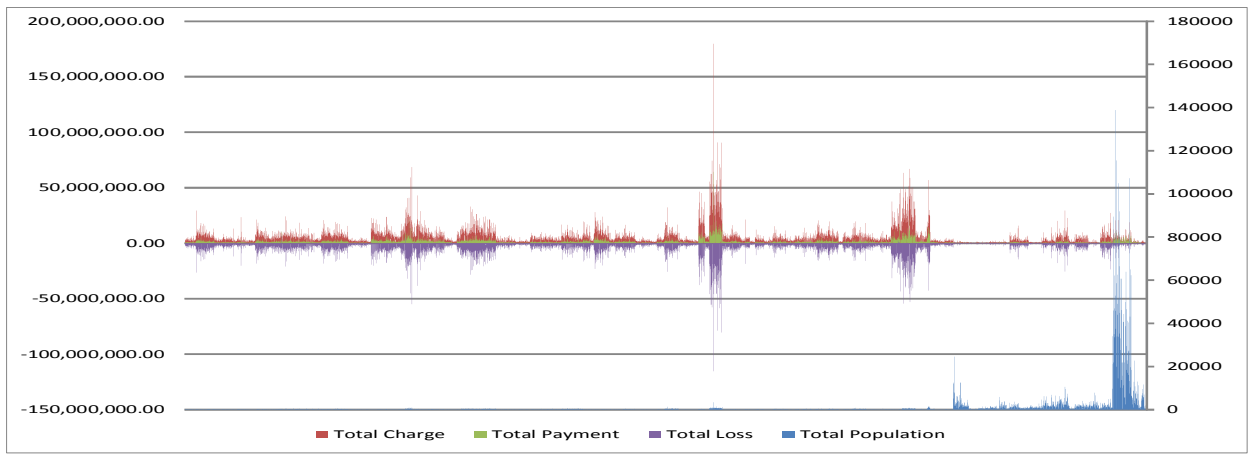

Here we see charges widely outstrip authorized payments, the difference (loss) is also varied across the program. Further, outpatient (far right) DRG's take on most of the people receiving services, yet garner lower charges and even lower payments than inpatient care (with variation).

The MTDS offers facility level transparency, at least for one health insurance system inside the United States. One of many government health care funding schemes, a full understanding of health transactions would require line item surveillance of private, public and personal expenditures on health. While the relationship between charge and undocumented cost may be a data mining nightmare there is sufficient data reported in the MTDS to evaluate the added contribution each class value has on the annualized programmatic structure of the data set when an Artificial Intelligence method like Naïve Bayes is used to discover and detail structures within the MTDS. This publication explores the structural dependencies of each class (variable) of the 2011 MTDS.

## Methods

The Naïve Bayes classifier has many uses and deployments[viii]. Over the last several years it has become the method of choice and a popular resource for the innovation and testing of data mining and predictive methods[ix]. Using a traditional posterior distribution model, Naïve efforts assume the independent of information in a data set and asks how this independent evidence fits with other variables and then returns a predictive value. Traditional Artificial Intelligence schema such as training values, uncertainty matrices and cross (annotated: X) validation are methodologically relevant but too complex to be detailed here. Contrary to popular convention, the 'independence' within a Naïve model is not necessary for the test to work (it is a pun or a play on words) but rather it is the variation of that independence (which cannot be total for a Bayesian statistician) from other evidence or data that rates and offers relative predictive value, training opportunities, accuracy and various contributions to statistical outputs in a Naïve Bayes operator.

This paper evaluates the MTDS for the independence (increase or decrease in accuracy and Kappa scores) of each class (variable) contained within the data set in a blinded framework using the Rapid Miner Open Source Data Mining Suite. The use of a Naïve evaluation on financial data is fairly rare outside of fraud detection and even more novel for an annualized evaluation of a programmatic outcomes data set. Due to discretization[x], the availability of time series data and other challenges of working with transaction data AI methods are often avoided in favor of health econometric and patient level evaluation. The use of a data mining method like NAÏVE Bayes on a data set like this requires some grounding in both transaction data, data structures and blind testing. Below I detail briefly, the overall concepts as well as the study protocol.

## The Data Set

The Data Model: The data follows a traditional rectangular format (A spreadsheet uses this format) where variable names are printed in the left side of the tuple and each transaction is detailed below in alphabetical order by provider facility. Transactions are detailed by individual provider aggregated charges by number discharged or the total DRGs by transaction type .

Blind Testing: Blind Redaction is a data mining method. It is used to detect the added value of a class\variable to the overall structure of a data set. It can be used on any rectangular data matrix. Some considerations are necessary for spatial, square and triangular matrices. It detects the added value of a class\variable by subtracting - redacting a whole class of information from the data set and applying a single statistical predictability or learning test for each class value. The redacted variable is then returned to the data set and another variable is redacted and the remaining values are retested until we have a statistical score for the predicative evidence of each variable under each blinded condition, one unique data set and one statistical test per variable class within the original data set. The scores are then graphed using a simple method (line, bar, box plot) and displayed to quickly discern the collapse of the accuracy of a prediction of a class value given the absence of a blind class. This value (of the accuracy of the data in the face of the blinded class) allows us to see the added value (given the blinded-subtraction) its absence would have lent to the prediction of a given class. A Naïve Bayes classifier gets extra statistical umph* from running the model and training it to predict classes multiple times (this is called X (Pronounced "cross") validation as it is typically, though not necessarily run ten times). The averages of the accuracy scores given X validation is the MICRO score which is the average of the accuracy out of 100%. This protocol ran the cross validation 10 times.

Protocol: The order of operations is to compile the data, transform through aggregation and discharge adjustment of discharge averaged transactions, label the classes, blind the data (iteratively), compute the score using Naïve Bayes, apply cross x (10) validation to the Naïve model, and record the statistical scores; and then repeat until each score is populated (see tables 1 and 2). Finally we graph the iterative scores to see the added value an absence (blind subtraction) in the data makes. Kappa and Accuracy scores were used as the statistical output in this model. MIKRO scores or the average of the model's Kappa and accuracy were also recorded.

Results

If accuracy is low (near 0), the available data did not inform the predictability of the value. If the accuracy is high (near 100) then the component data variables in the data set (that was not blinded) enable predictability of the accuracy value proportionately. The Kappa shows the agreement between the models during X validation. The MIKRO, or average accuracy or average Kappa will show us how the models fared when X trained to predict the best model given the available data. The Accuracy or Kappa + is the most accurate model\agreement out of ten X training sessions, - is the lowest (often the first) and MIKRO is the average of all trained sets under the X validation operation. Please note that graph 2 and graph 4 have the same data, they are simply sorted differently to demonstrate dimensionality; the same is true of graphs 3 and 5.

Table One: Accuracy MIKRO Scores

| | Accuracy | Charge per Discharge | Facility Total Charge | Discharges | DRG | Loss per Discharge | Facility Total Loss | Payment per Discharge | Facility Total Payment | State | ZIP |
|---|---|---|---|---|---|---|---|---|---|---|---|
| BLIND | Charge per Discharge | N\A | 99.28 | 99.96 | 1.23 | 97.96 | 98.66 | 95.98 | 99.42 | 99.69 | 99.83 |
| BLIND | Facility Total Charge | 98.27 | N\A | 99.84 | 1.23 | 98.12 | 98.37 | 95.82 | 99.52 | 99.74 | 99.83 |
| BLIND | CITY | 98 | 98.93 | 99.75 | 1.89 | 97.89 | 98.14 | 95.83 | 99.21 | 99.37 | 99.57 |
| BLIND | DISCHARGE | 98.13 | 99.09 | N\A | 1.14 | 97.99 | 98.41 | 96.09 | 99.33 | 99.75 | 99.81 |
| BLIND | DRG | 98.16 | 98.98 | 99.84 | N\A | 97.88 | 98.26 | 94.22 | 99.23 | 99.66 | 99.78 |
| BLIND | INorOUTPT | 98.24 | 99.23 | 99.93 | 1.06 | 98.05 | 98.6 | 95.65 | 99.46 | 99.66 | 99.77 |
| BLIND | Loss per Discharge | 97.91 | 99.27 | 99.96 | 1.21 | N\A | 98.66 | 95.95 | 99.43 | 99.69 | 99.83 |
| BLIND | Facility total Loss | 98.29 | 99.13 | 99.89 | 1.22 | 98.13 | N\A | 95.83 | 99.54 | 99.74 | 99.83 |
| BLIND | Payment per Discharge | 91.1 | 99.3 | 99.95 | 0.1 | 97.92 | 98.72 | N\A | 99.52 | 99.78 | 99.86 |
| BLIND | Facility Total Paymemt | 98.22 | 99.25 | 99.62 | 1.2 | 98.06 | 98.06 | 95.68 | N\A | 99.71 | 99.8 |
| BLIND | PROVIDER | 99.38 | 98.89 | 99.73 | 6.42 | 97.72 | 98.02 | 95.72 | 99.2 | 99.2 | 99.2 |
| BLIND | STATE | 98.25 | 99.06 | 99.91 | 1.3 | 98.1 | 98.39 | 95.75 | 99.26 | N\A | 99.19 |
| BLIND | ZIP | 98.21 | 99.1 | 99.9 | 1.3 | 98.07 | 98.4 | 95.75 | 99.35 | 98.81 | N\A |

When Payment per Discharge is blind, Charge per Discharge becomes 7% harder to predict across model values. DRG is comparatively unlearnable with an average of 6% being the highest accuracy when individual provider is blinded from the model. Other predictions were between 94 and 99% accurate, or predictable given the other data variables in the model.

Table Two: Kappa MIKRO Scores

| | Kappa | Charge per Discharge | Facility Total Charge | Discharges | DRG | Loss per Discharge | Facility Total Loss | Payment per Discharge | Facility Total Paymemt | State | ZIP |
|---|---|---|---|---|---|---|---|---|---|---|---|
| BLIND | Charge per Discharge | N\A | 0.095 | 0.595 | 0.005 | 0.386 | 0.172 | 0.487 | 0.038 | 0.997 | 0.998 |
| BLIND | Facility Total Charge | 0.423 | N\A | 0.289 | 0.005 | 0.427 | 0.119 | 0.481 | 0.038 | 0.997 | 0.998 |
| BLIND | CITY | 0.392 | 0.069 | 0.222 | 0.011 | 0.403 | 0.131 | 0.483 | 0.028 | 0.993 | 0.995 |
| BLIND | DISCHARGE | 0.398 | 0.077 | N\A | 0.004 | 0.407 | 0.149 | 0.481 | 0.033 | 0.997 | 0.998 |
| BLIND | DRG | 0.404 | 0.07 | 0.267 | N\A | 0.394 | 0.137 | 0.377 | 0.029 | 0.996 | 0.997 |
| BLIND | INorOUTPT | 0.406 | 0.087 | 0.437 | 0.003 | 0.406 | 0.163 | 0.461 | 0.041 | 0.996 | 0.997 |
| BLIND | Loss per Discharge | 0.351 | 0.094 | 0.601 | 0.004 | N\A | 0.172 | 0.481 | 0.039 | 0.997 | 0.998 |
| BLIND | Facility total Loss | 0.425 | 0.074 | 0.364 | 0.004 | 0.428 | N\A | 0.481 | 0.047 | 0.997 | 0.998 |
| BLIND | Payment per Discharge | 0 | 0.095 | 0.484 | -0.006 | 0.377 | 0.175 | N\A | 0.045 | 0.998 | 0.998 |
| BLIND | Facility Total Paymemt | 0.416 | 0.091 | 0.119 | 0.004 | 0.42 | 0.42 | 0.469 | N\A | 0.997 | 0.997 |
| BLIND | PROVIDER | 0.067 | 0.067 | 0.207 | 0.057 | 0.384 | 0.126 | 0.479 | 0.028 | 0.992 | 0.99 |
| BLIND | STATE | 0.42 | 0.075 | 0.442 | 0.005 | 0.426 | 0.147 | 0.472 | 0.03 | N\A | 0.99 |
| BLIND | ZIP | 0.415 | 0.079 | 0.389 | 0.005 | 0.422 | 0.148 | 0.474 | 0.034 | 0.988 | N\A |

Facility total Charge, Facility total Loss and Facility total Payment averaged low kappa for agreement between models under X validation. This indicates that these fields are the hardest to learn and predict given the available un-blinded data in the set. When Payment per Discharge is redacted the model lost integrity (learned the wrong things) when attempting to predict DRG.

### Graph Two: Effects of Values (top) when Blinded(bottom) Accuracy Range

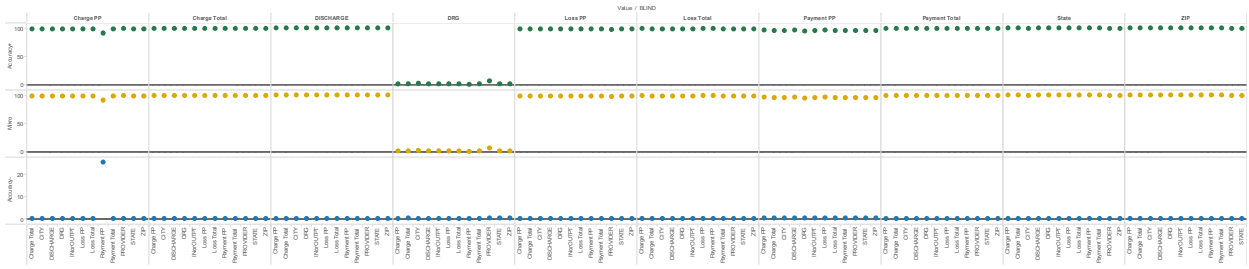

Highest accuracy approaches 100% for the best (Most learned) X validated models. DRG is notably unpredictable with even the average training score being very low. Further, Payment PP (pp= per discharge\ per person) and Charge PP blinded by Payment PP has lower learning success.

### Graph Three: Effects of Values (Top) when Blinded(Bottom) Kappa Range

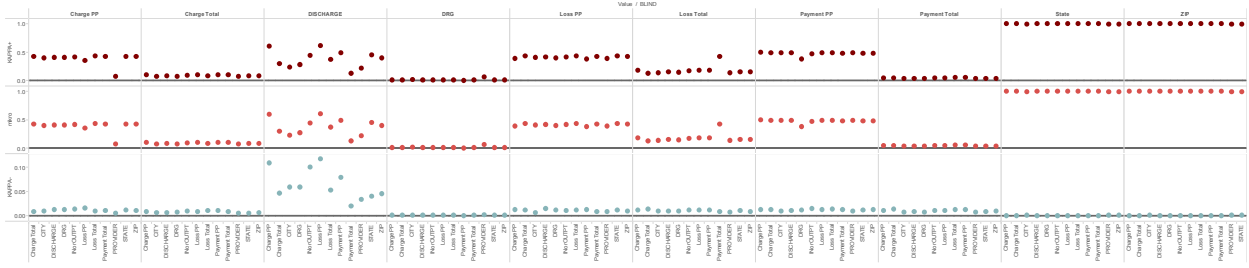

Model agreement found four patterns, Zip and State show weak openings, and strong finishes from learning. Payment Total, DRG, and Charge Total show weak agreement between models which does not improve. Charge PP, loss PP, and Payment PP follow a modest agreement improvement. Lastly the inconsistent agreement for discharge models suggests that the values in discharge class are highly dependent on other values when achieving model agreement.

### Graph Four: Effects of Values (Bottom) when Blinded (Top) Accuracy

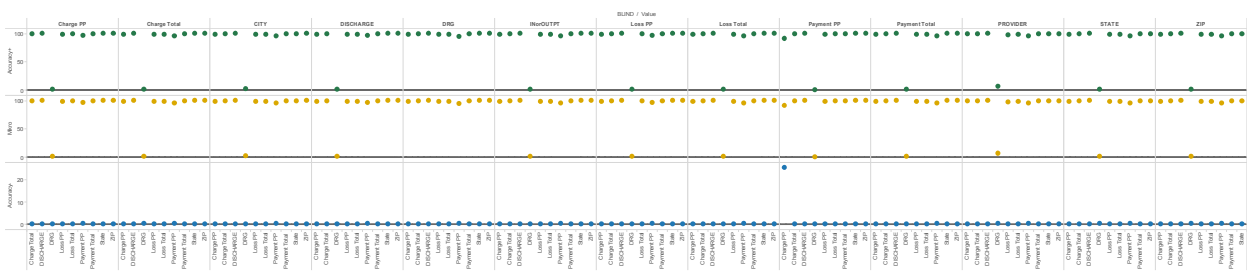

When Payment PP is Blinded, Charge pp becomes readily predictable as the lowest accuracy score approaches 100%. Predicting DRG does not have the same relationship even after cross validation sessions.

Graph Five: **Effects of Values (Bottom) when Blinded (Top) Kappa**

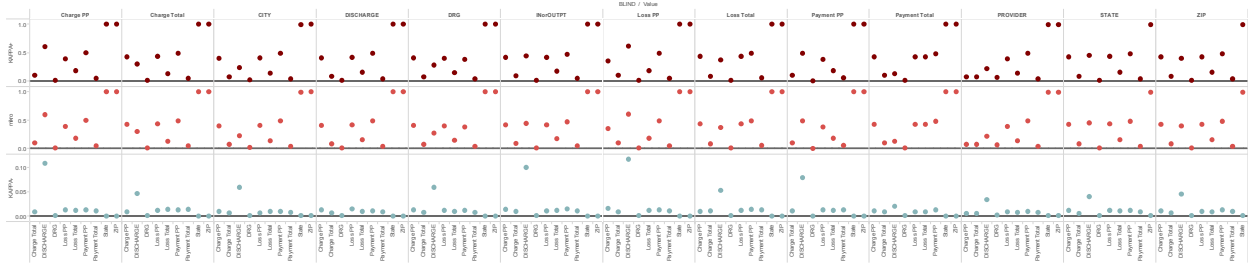

The Kappa of the learning process is fairly consistent here, as values respond differently (have different added value) given the blinded sets. Further many Kappa's are consistent across the blinded sets suggesting good model cohesion and learning ability. Discharge appears easier to learn initially than other elements.

Discussion: Medicare is the payer of last resort. This position within a wider insurance scheme should be considered when evaluating Medicare transactions as the specific procedure is not predictable given the financial constraints and the availability of strong prediction values of the same legality. While discretization by a financial range may help to classify DRG values (assuming such classes are uniform across providers, geographies and price charts), in and out patient difference and cost differences between in and outpatient procedures failed to train the models appropriately to predict DRG. Further the data model of the transaction set itself is not sufficient evidence of behavior to predict DRG despite the transaction values contained in the set and strong prediction values from other model runs.

Conclusions

The overlay hypotheses such as charges being driven by the volume served or DRG being related to charges or payments is readily false in this analysis despite 28 million Americans being billed through Medicare in 2011 and the program distributing over 70 billion in this transaction set alone. It may be impossible to predict the dependencies and data structures of the payer of last resort without data from payers of first and second resort. Political concerns about Medicare would be better served focusing on these first and second order payer systems as what Medicare costs is not dependent on Medicare itself.